\documentclass{article} 
\usepackage[table]{xcolor}
\usepackage{iclr2016_conference,times}
\usepackage{hyperref}
\usepackage{url}

\usepackage{algorithm}
\usepackage{algorithmic}
\usepackage{amsmath}

\usepackage{subcaption}
\usepackage{graphicx} 
\usepackage{footnote}
\usepackage{amsfonts}
\usepackage{textcomp}

\newcolumntype{P}[1]{>{\centering\arraybackslash}p{#1}}

\usepackage{comment}
\usepackage{multirow}
\usepackage{soul}

\usepackage{slashbox}
\usepackage{makecell}
\usepackage{hyperref}

\usepackage{tabularx}
\usepackage{makecell}

\usepackage{diagbox}

\usepackage{tikz}
\definecolor{color3}{rgb}{0.000, 1.000, 0.000}
\definecolor{color4}{rgb}{0.459, 0.937, 0.000}
\definecolor{color5}{rgb}{0.604, 0.871, 0.000}
\definecolor{color6}{rgb}{0.710, 0.796, 0.000}
\definecolor{color7}{rgb}{0.796, 0.710, 0.000}
\definecolor{color8}{rgb}{0.871, 0.604, 0.000}
\definecolor{color9}{rgb}{0.937, 0.459, 0.000}
\definecolor{color10}{rgb}{1.000, 0.000, 0.000}
\definecolor{color_10_0}{rgb}{0.000, 1.000, 0.000}
\definecolor{color_10_1}{rgb}{0.396, 0.957, 0.000}
\definecolor{color_10_2}{rgb}{0.522, 0.914, 0.000}
\definecolor{color_10_3}{rgb}{0.616, 0.867, 0.000}
\definecolor{color_10_4}{rgb}{0.690, 0.812, 0.000}
\definecolor{color_10_5}{rgb}{0.757, 0.757, 0.000}
\definecolor{color_10_6}{rgb}{0.812, 0.690, 0.000}
\definecolor{color_10_7}{rgb}{0.867, 0.616, 0.000}
\definecolor{color_10_8}{rgb}{0.914, 0.522, 0.000}
\definecolor{color_10_9}{rgb}{0.957, 0.396, 0.000}
\definecolor{color_10_10}{rgb}{1.000, 0.000, 0.000}

\usetikzlibrary{arrows,automata}

\usepackage{array}
\newcolumntype{C}[1]{>{\centering\arraybackslash}p{#1}}

\newcolumntype{C}{ >{\centering\arraybackslash} m{1cm} }
\newcolumntype{D}{ >{\centering\arraybackslash} m{3cm} }

\definecolor{LimeGreen}{rgb}{0.1,0.9,0.1}
\definecolor{Maroon}{rgb}{0.9,0.1,0.1}
\definecolor{bananayellow}{rgb}{1.0, 0.88, 0.21}
\definecolor{cadmiumyellow}{rgb}{1.0, 0.96, 0.0}

\def\@fnsymbol#1{\ensuremath{\ifcase#1\or \daggerdagger\or \ddagger\or
   \mathsection\or \mathparagraph\or \|\or **\or \dagger
   \or \ddagger\ddagger \else\@ctrerr\fi}}

\let\oldcite\cite
\renewcommand{\cite}[1]{[\oldcite{#1}]}

\newcommand{\no}{ $\color{Maroon}\times$ }
\newcommand{\yes}{ $\color{LimeGreen}\checkmark$  }

\newcommand\mborder[2][]{%
  \tikz[anchor=base,baseline]{
    \node[inner sep=1pt,#1](h){$\displaystyle#2\mathstrut$};
    \draw(h.south east)--(h.south west)--(h.north west)
       --(h.north east)--(h.south east);
  }%
}

\newcommand{\argmax}{\arg\!\max}

\newcommand{\eqn}[1]{Eqn.~\ref{eqn:#1}}
\newcommand{\fig}[1]{Fig.~\ref{fig:#1}}
\newcommand{\tab}[1]{Table~\ref{tab:#1}}
\newcommand{\secc}[1]{Section~\ref{sec:#1}}

\usepackage[font=small,tableposition=top]{caption}
\DeclareCaptionLabelFormat{andtable}{#1~#2  \&  \tablename~\thetable}

\title{Learning Simple Algorithms from Examples}

\author{Wojciech Zaremba$^{\S}$\\
New York University\\
\texttt{woj.zaremba@gmail.com} \\
\And
Tomas Mikolov \qquad Armand Joulin \qquad Rob Fergus\\
\qquad \qquad \qquad Facebook AI Research \\ 
\texttt{\{tmikolov,ajoulin,robfergus\}@fb.com} \\
}

\begin{document}

\maketitle

\vspace{-3mm}
\begin{abstract}
  We present an approach for learning simple algorithms such as
  copying, multi-digit addition and single digit multiplication
  directly from examples. Our framework consists of a set of
  \textit{interfaces}, accessed by a \textit{controller}. Typical
  interfaces are 1-D tapes or 2-D grids that hold the input and output
  data.  For the controller, we explore a range of neural
  network-based models which vary in their ability to abstract the
  underlying algorithm from training instances and  generalize to test examples with many thousands of digits.  The controller is
  trained using $Q$-learning with several enhancements and we show
  that the bottleneck is in the capabilities of the controller rather
  than in the search incurred by $Q$-learning.
\end{abstract}
\footnotetext[4]{Work done while the author was at Facebook AI Research.}

\vspace{-7mm}
\section{Introduction}
\vspace{-3mm}
Many every day tasks require a multi-step interaction with the world. For
example, picking an apple from a tree requires visual localization
of the apple; extending the arm and then fine muscle control, guided
by visual feedback, to pluck it
from the tree. While each individual procedure is not complex, the
task nevertheless requires careful sequencing of operations across both visual and motor systems.  

This paper explores how machines can learn algorithms involving a similar
compositional structure. Since our emphasis is on learning the correct
sequence of operations, we consider the domain of arithmetic where the
operations themselves are very simple. For example, although learning
to add two digits is straightforward, solving addition of two
multi-digit numbers requires precise coordination of this operation
with movement over the sequence and recording of the carry. We explore
a variety of algorithms in this domain, including complex tasks
involving addition and multiplication.

Our approach formalizes the notion of a central controller that
interacts with the world via a set of interfaces, appropriate to the
task at hand. The controller is a neural network model which must learn to
control the interfaces, via a set of discrete actions (e.g.~``move
input tape left'', ``read'', ``write symbol to output tape'', ``write
nothing this time step'' ) to produce the correct output for given
input patterns. Specifically, we train the controller from large sets
of examples of input and output patterns using reinforcement
learning. Our reward signal is sparse, only being received when the
model emits the correct symbol on the output tape.

 We consider two separate settings. In the first, we provide supervision in
 the form of ground truth actions. In the second, we train only with input-output pairs (i.e.~no supervision over actions). 
 While we are able to solve all the tasks in the latter case, the supervised setting provides insights about the model limitations and an upper bound on
 the performance. 
We evaluate our model on sequences far longer than those present
during training. Surprisingly, we find that controllers with even
modest capacity to recall previous states can easily overfit the short
training sequences and not generalize to the test examples, even if
the correct actions are provided. 
Even with an appropriate controller, off-the-shelf $Q$-learning fails on
the majority of our tasks. We therefore introduce a series of
modifications that dramatically improve performance. These include:
(i) a novel
dynamic discount term that makes the reward invariant to the sequence
length; (ii) an extra penalty that aids generalization
and (iii) the deployment of Watkins Q-lambda \cite{sutton1998reinforcement}. 

We would like to direct the reader to the video accompanying this
paper
(\href{https://youtu.be/GVe6kfJnRAw}{\color{blue}{https://youtu.be/GVe6kfJnRAw}}). This
gives a concise overview of our approach and complements the following
explanations. Full source code for this work can be found at
\href{https://github.com/wojzaremba/algorithm-learning}{\color{blue}
  {https://github.com/wojzaremba/algorithm-learning}}.

\section{Model}
Our model consists of an RNN-based controller that accesses the
environment through a series of pre-defined interfaces. Each interface has a
specific structure and set of actions it can perform. The interfaces
are manually selected according to the task (see
\secc{tasks}). The controller is the only part of the system that
learns and has no prior knowledge of how the
interfaces operate. Thus the controller must learn the sequence of
actions over the various interfaces that allow it to solve a
task. We make use of three different interfaces:

\noindent {\bf Input Tape:} This provides access to the input data
symbols stored on an ``infinite'' 1-D tape. A read head accesses a single character
at a time through the \textit{read} action. The head can be moved via
the \textit{left} and \textit{right} actions.
 
\noindent {\bf Input Grid:} This is a 2D version of the input
tape where the read head can now be moved by actions  \textit{up},
\textit{down}, \textit{left} and \textit{right}. 

\noindent {\bf Output Tape:} This is similar to the input tape,
except that the head now writes a single symbol at a time to the tape,
as provided the controller. The vocabulary includes a no-operation
symbol (NOP) enabling the controller to defer output if it
desires. During training, the written and target symbols are compared
using a cross-entropy loss. This provides a differentiable learning
signal that is used in addition to the sparse reward signal provided by the
$Q$-learning.

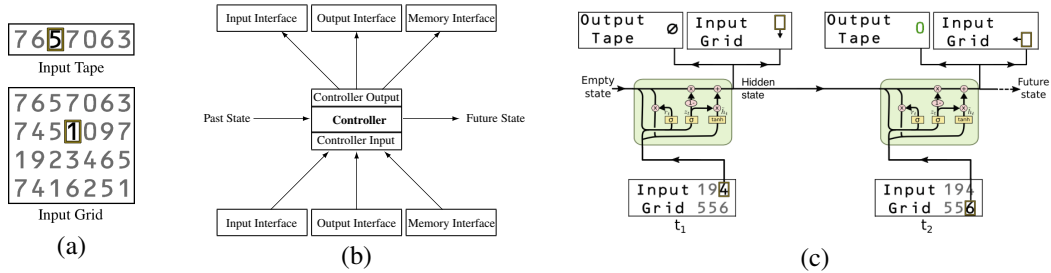
\begin{figure}
  \begin{center}
    \begin{minipage}{0.12\linewidth}
      \centering
      \includegraphics[width=\linewidth]{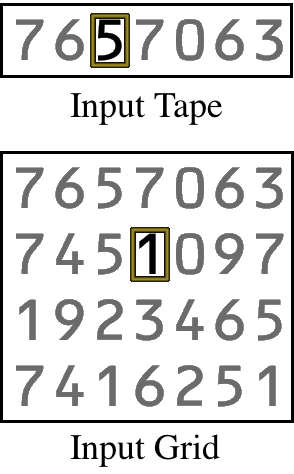}
      \\
      (a)
    \end{minipage}
    \begin{minipage}{0.41\linewidth}
      \centering
      \scalebox{0.65}{
        \begin{picture}(200, 130)
          \put(74, 60){\framebox(52, 15){}}
          \put(86, 65){{\scriptsize \textbf{Controller}}}
          \put(74, 50){\framebox(52, 10){}}
          \put(77, 53){{\scriptsize Controller Input}}
          \put(74, 75){\framebox(52, 10){}}
          \put(75, 78){{\scriptsize Controller Output}}

          \put(50, 15){\vector(1, 1){34}}
          \put(19, 0){\framebox(52, 15){}}
          \put(24, 5){{\scriptsize Input Interface}}

          \put(100, 15){\vector(0, 1){34}}
          \put(74, 0){\framebox(52, 15){}}
          \put(77, 5){{\scriptsize Output Interface}}

          \put(150, 15){\vector(-1, 1){34}}
          \put(129, 0){\framebox(52, 15){}}
          \put(130, 5){{\scriptsize Memory Interface}}

          \put(90, 85){\vector(-1, 1){34}}
          \put(19, 119){\framebox(52, 15){}}
          \put(24, 124){{\scriptsize Input Interface}}

         \put(100, 85){\vector(0, 1){34}}
          \put(74, 119){\framebox(52, 15){}}
          \put(77, 124){{\scriptsize Output Interface}}

          \put(115, 85){\vector(1, 1){34}}
          \put(129, 119){\framebox(52, 15){}}
          \put(130, 124){{\scriptsize Memory Interface}}

          \put(40, 68){\vector(1, 0){33}}
          \put(127, 68){\vector(1, 0){33}}
          \put(11, 66){{\scriptsize Past State}}
          \put(164, 66){{\scriptsize Future State}}

        \end{picture}
      }
      \\
      (b)
    \end{minipage}
    \begin{minipage}{0.45\linewidth}
      \centering
      \includegraphics[width=\linewidth]{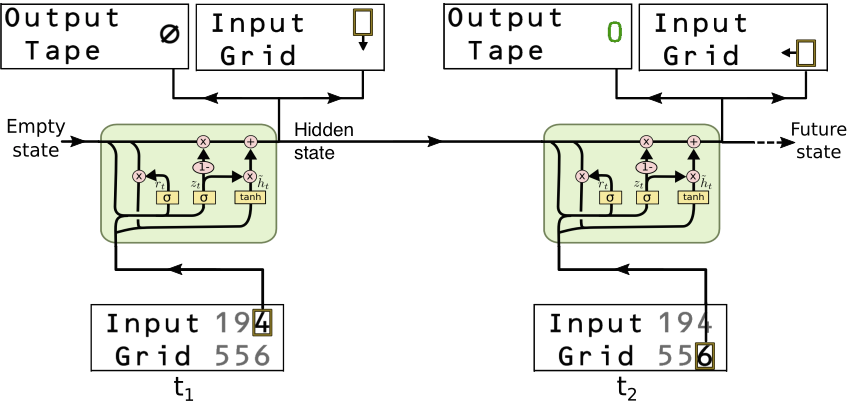}
      \\
      (c)
    \end{minipage}
  \end{center}
  \caption{(a): The input tape and grid interfaces. Both have a single
    head (gray box) that reads one character at a time, in response to
    a \textit{read} action from the controller. It can also move the
    location of the head with the \textit{left} and \textit{right} (and \textit{up}, \textit{down}) actions. (b) An
    overview of the model, showing the abstraction of controller and a
    set of interfaces (in our experiments the memory interface is not used). (c) An example of the model applied to the
    addition task. At time step $t_1$, the controller, a form of
    RNN, reads the symbol $4$ from the input grid and outputs a no-operation
    symbol ($\oslash$) on the
    output tape and a \textit{down} action on the input interface, as well
    as passing the hidden state to the next timestep.}
  \label{fig:interfaces}
\end{figure}

\fig{interfaces}(a) shows examples of the input tape and
grid interfaces. \fig{interfaces}(b) gives an overview of our
controller--interface abstraction and \fig{interfaces}(c) shows an
example of this on the addition task (for two time steps).

For the controller, we explore several recurrent neural network
architectures: two different sizes of 1-layer LSTM \cite{hochreiter1997long},
a gated-recurrent unit (GRU)\cite{cho2014learning} and a vanilla
feed-forward network. Note that RNN-based models are able to
remember previous network state, unlike the the feed-forward
network. This is important because some tasks explicitly require
some form of memory, e.g. the carry in addition.  

The simple algorithms we consider (see \secc{tasks}) have
deterministic solutions that can be expressed as a finite state
automata. Thus during training we hope the controller will implicitly
learn the correct automata from the training samples, since this would
ensure generalization to sequences of arbitrary length. 
On some tasks like reverse, we observe a higher-order form of over-fitting: the
model learns to solve the training tasks correctly and generalizes
successfully to test sequences of the same length (thus is not
over-fitting in the standard sense). However, when presented
with longer test sequences the model fails completely. This
suggests that the model has converged to an incorrect local minima,
one corresponding to an alternate automata which have an implicit
awareness of the sequence length of which they were trained. See \fig{automata} for an example of this on the reverse
task. Note that this behavior results from the controller, not the
learning scheme, since it is present in both the supervised
(\secc{supervised}) and $Q$-learning settings (\secc{qlearn}). These
experiments show the need to carefully adjust the controller capacity
to prevent it learning any dependencies on the length
of training sequences, yet ensuring it has enough state to implement
the algorithm in question.

As illustrated in \fig{interfaces}(c), the controller passes two
signals to the output tape: a discrete action (move left, move right,
write something) and a symbol from the vocabulary. This symbol is
produced by taking the max from the softmax output on the top of the
controller. In training, two different signals are computed from this: (i) a
cross-entropy loss is used to compare the softmax output to the
target symbol and (ii) a discrete 1/0 reward if the symbol is
correct/incorrect. The first signal gives a continuous gradient to
update the controller parameters via backpropagation. Leveraging the
reward requires reinforcement learning, since many actions might occur
before a symbol is written to the output tape. Thus the action output
of the controller is trained with reinforcement learning and the
symbol output is trained by backpropagation. 

\section{Tasks}
\label{sec:tasks}

\begin{figure}[h!]
  \centering
  \fbox{
    \begin{minipage}[t]{0.14\linewidth}
      \centering
      \vspace{2mm}
      \includegraphics[height=0.5\linewidth]{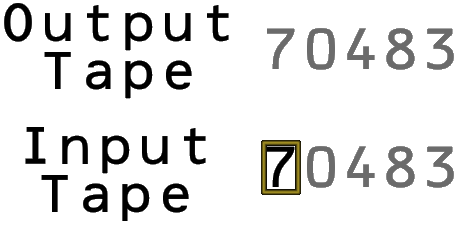}
      \vspace{2mm}
      {\small{\tiny\\}Copy}
      \vspace{3mm}
    \end{minipage}
  }
  \fbox{
    \begin{minipage}[t]{0.14\linewidth}
      \vspace{2mm}
      \centering
      \includegraphics[height=0.5\linewidth]{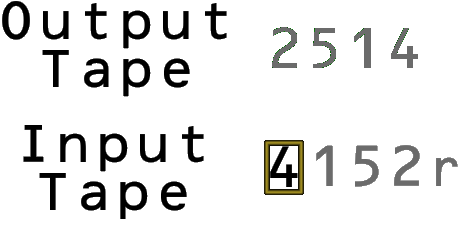}
      \vspace{2.5mm}
      {\small{\tiny\\}Reverse}
      \vspace{3mm}
    \end{minipage}
  }
  \fbox{
    \begin{minipage}[t]{0.14\linewidth}
      \vspace{2mm}
      \centering
      \includegraphics[height=0.6\linewidth]{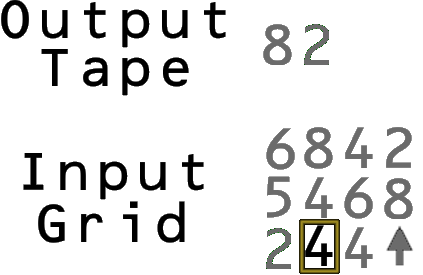}
      \vspace{3mm}
      {\small{\tiny\\}Walk}
      \vspace{3mm}
    \end{minipage}
  }
  \fbox{
    \begin{minipage}[t]{0.14\linewidth}
      \vspace{2mm}
      \centering
      \includegraphics[height=0.5\linewidth]{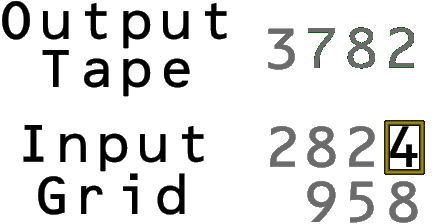}
      \vspace{2.5mm}
      {\small{\tiny\\}Addition}
      \vspace{3mm}
    \end{minipage}
  }
  \fbox{
    \begin{minipage}[t]{0.14\linewidth}
      \vspace{2mm}
      \centering
      \includegraphics[height=0.6\linewidth]{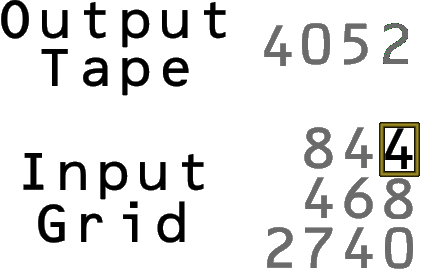}
      \vspace{1mm}
      {\small{\tiny\\}3~number\\addition}
      \vspace{1mm}
    \end{minipage}
  }
  \fbox{
    \begin{minipage}[t]{0.14\linewidth}
      \vspace{2mm}
      \centering
      \includegraphics[height=0.5\linewidth]{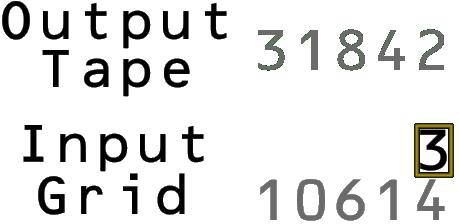}
      \vspace{0.5mm}
      {\small{\tiny\\}Single digit\\multiplication}
      \vspace{1mm}
    \end{minipage}
  }
  \caption{Examples of the six tasks, presented in their initial state. The yellow box indicates the
    starting position of the read head on the Input Interface. The
    gray characters on the Output Tape are \textbf{target}
    symbols used in training. }
  \label{fig:tasks}
\end{figure}

We consider six different tasks: copy, reverse, walk, multi-digit addition, 3
number addition and single digit multiplication. The input interface
for copy and reverse is an input tape, but an input grid for the
others. All tasks use an output tape interface. Unless otherwise
stated, all arithmetic operations use base 10. Examples of the six
tasks are shown in \fig{tasks}. 

\noindent {\bf Copy:} This task involves copying the symbols from the
input tape to the output tape. Although simple, the model still has to
learn the correspondence between input and output symbols, as well as
executing the move right action on the input tape.

\noindent {\bf Reverse:} Here the goal is to reverse a sequence of
symbols on the input tape. We provide a special character ``r'' to
indicate the end of the sequence. The model must learn to move right
multiple times until it hits the ``r'' symbol, then move to the left,
copying the symbols to the output tape. 

\noindent {\bf Walk:} The goal is to copy symbols, according to the
directions given by an arrow symbol.  The controller starts by moving
to the right (suppressing prediction) until reaching one of the
symbols $\uparrow, \downarrow, \leftarrow$.  Then it should change
it's direction accordingly, and copy all symbols encountered to the
output tape.

\noindent {\bf Addition:} The goal is to add two multi-digit
sequences, provided on an input grid.  The sequences are provided in
two adjacent rows, with their right edges aligned. The initial
position of the read head is the last digit of the top number
(i.e.~upper-right corner). The model has to: (i) memorize an addition
table for pairs of digits; (ii) learn how to move over the input grid
and (iii) discover the concept of a carry.

\noindent {\bf 3 Number Addition:}  As for the addition task, but now
three numbers are to be added. This is more challenging as the reward
signal is less frequent (since more correct actions must be completed before
a correct output digit can be produced). Also the carry now can take
on three states (0, 1 and 2), compared with two for the 2 number
addition task. 
  
\noindent {\bf Single Digit Multiplication:} This involves multiplying
a single digit with a long multi-digit number. It is of similar
complexity to the 2 number addition task, except that the carry can
take on more values $\in [0, 8]$.

\section{Related Work}
\label{sec:related}

A variety of recent work has explored the learning of simple
algorithms. Many of them are different embodiments of the
controller-interface abstraction formalized in our model.  The Neural
Turing Machine (NTM) \cite{graves2014neural} uses a modified LSTM
\cite{hochreiter1997long} as the controller, and has three inferences: sequential input,
delayed output and a differentiable memory. The model is able to learn
simple algorithms including copying and sorting.  The Stack RNN
\cite{joulin2015inferring} has an RNN controller and three interfaces:
sequential input, a stack memory and sequential output. The learning
of simple binary
patterns and regular expressions is demonstrated. A closely related
work to this is \cite{Das92}, which was recently extended in the Neural DeQue
\cite{grefenstette2015learning} to use a list instead. End-to-End Memory Networks
\cite{sukhbaatar2015weakly} use a feed-forward network as the
controller and interfaces consisting of a soft-attention input, plus a
delayed output (by a fixed number of ``hops''). The model is applied to simple Q\&A tasks, some of
which involve logical reasoning. In contrast, our model automatically
determines when to produce output and uses more general
interfaces. 

However, most of these approaches use continuous interfaces that permit
training via back-propagation of gradients. Our approach differs in
that it uses discrete interfaces thus is more challenging to train
since as we must rely on reinforcement learning instead. A notable
exception is the Reinforcement Learning Neural Turing Machine (RLNTM)
\cite{zaremba2015reinforcement} which is a version of the NTM with
discrete interfaces. The Stack-RNN \cite{joulin2015inferring} also
uses a discrete search procedure for its interfaces but it is unclear
how this would scale to larger problems. 
 
The problem of learning algorithms has its origins in the field of
program induction \cite{nordin1997evolutionary, liang2013learning,
  wineberg1994representation,solomonoff1964formal}. In this domain, the model has to infer
the source code of a program that solves a given problem.  This is a 
similar goal to ours, but in quite a different setting. I.e.~we do
not produce a computer program, but rather a neural net that can
operate with interfaces such as tapes and so implements the program
without being human-readable.
A more relevant work is \cite{schmidhuber2004optimal} which learns an algorithms for the Hanoi tower
problem, using a simple form of program induction and incremental
learning components. Genetic algorithms \cite{Holland,Goldberg} also
can be considered a form of program induction, but are mostly based on a
random search strategy rather than a learned one.

Similar to \cite{mnih2013playing}, we train the controller to
approximate the $Q$-function. However, we introduce several
modifications on top of the classical $Q$-learning. First, we 
use Watkins $Q(\lambda)$ \cite{watkinsq,sutton1998reinforcement}. This helps to
overcome a \textit{non-stationary} environment. We are unaware of any prior work that
uses Watkins $Q(\lambda)$ for this purpose. 
Second, we reparametrized $Q$ function, to become invariant to the sequence
length. Finally, we penalize $||Q(s, \bullet)||$, which might help to remove
positive bias \cite{hasselt2010double}.

\section{Supervised Experiments}
\label{sec:supervised}
To understand the behavior of our model and to provide an upper bound
on performance, we train our model in a supervised setting, i.e.~where
the ground truth actions are provided. Note that the controller must
still learn which symbol to output. But this now can be done purely with
backpropagation since the actions are known. 

To facilitate comparisons of difficulty between tasks, we use a common
measure of \textit{complexity}, corresponding to the number of time steps
required to solve each task (using the ground truth
actions\footnote{In practice, multiple solutions can exist (see
Appendix A), thus the measure is approximate.}). For
instance, a reserve task involving a sequence of length $10$ requires
$20$ time-steps ($10$ steps to move to the ``r'' and $10$ steps to move back to
the start). The conversion factors between sequence lengths and
complexity are as follows: copy=1; reverse=2; walk=1; addition=2; 3 row
addition=3 and single digit multiplication=1.

For each task, we train a separate model, starting with sequences of
complexity 6 and incrementing by 4 once it achieves 100\% accuracy on
held-out examples of the current length. Training stops once the model
successfully generalizes to examples of complexity 1000. Three
different cores for the controllers are explored: (i) a 200 unit, 1-layer LSTM; (iii) a 200 unit, 1-layer GRU model and (iii)
a 200 unit, 1-layer feed-forward network. An additional linear layer
is placed on top of these model that maps the hidden state to either
action for a given interface, or the target symbol.

\begin{figure}[h]
  \centering
   \includegraphics[width=1\linewidth]{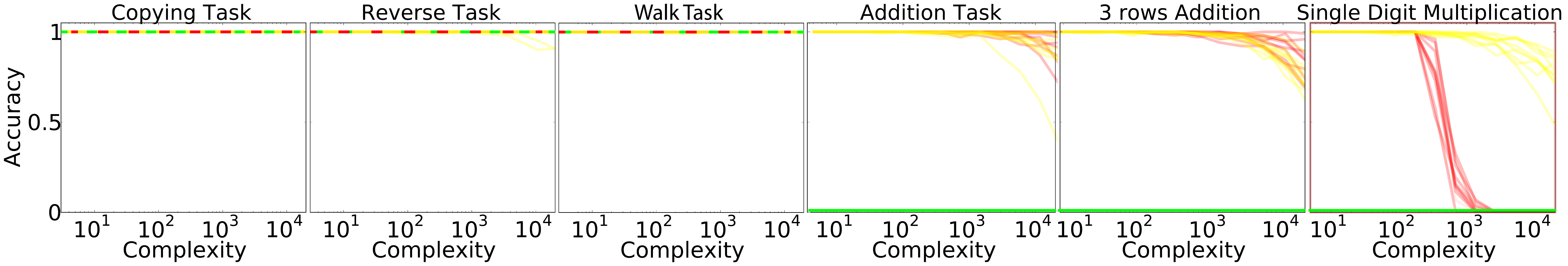}
 \caption{Test accuracy for all tasks with \textbf{supervised} actions
   over $10$ runs for feed-forward (green), GRU (red) and LSTM
   (yellow) controllers. In this setting the optimal policy is
   provided. Complexity is the number of time steps required to
   compute the solution. Every task has slightly different conversion 
   factor between complexity and the sequence length: a complexity of $10^4$ 
   for copy and walk would mean $10^4$ input symbols; for reverse would
   correspond to $\frac{10^4}{2}$ input symbols; 
   for addition would involve two $\frac{10^4}{2}$ long numbers; for 3 row
   addition would involve three $\frac{10^4}{3}$ long numbers and for
   single digit multiplication would involve a single $10^4$ long number.}
  \label{fig:supervised}
\end{figure}

In \fig{supervised} we show the accuracy of the different controllers
on the six tasks for test instances of increasing complexity, up to
$20,000$ time-steps. The
simple feed-forward controller generalizes perfectly on the copy,
reverse and walk tasks but completely fails on the remaining ones, due to a 
lack of required memory\footnote{Ammending the interfaces to allow
both reading and writing on the same interface would provide a mechanism for long-term memory, even with a
  feed-forward controller. But then the same lack of generalization issues
  (encountered with more powerful controllers) would become an issue.}. The RNN-based controllers succeed to varying
degrees, although some variability in performance is
observed. 

\begin{figure}[h]
  \begin{minipage}[c]{0.48\linewidth}
    \centering
   \vspace{10mm}
    \begin{tikzpicture}[>=stealth',shorten >=1pt,auto,node distance=2.1cm,scale=0.5, every node/.style={scale=0.55}]
      \node[state] (A)      {{right}};
      \node[state]         (B) [right of=A]  {{left}};
      \path[->] (A)  edge [loop above] node {} (A)
                     edge              node {} (B) 
                (B)  edge [loop above] node {} (B);
    \end{tikzpicture} \\
(a)
  \end{minipage}
  \hfill
  \begin{minipage}[c]{0.48\linewidth}
    \centering
    \begin{tikzpicture}[>=stealth',shorten >=1pt,auto,node distance=2.1cm,scale=0.5, every node/.style={scale=0.55}]
      \node[state] (A0)                {{right{\Large $_1$}}};
      \node[state]           (A1) [right of=A0]  {{right{\Large $_2$}}};
      \node[state]           (A2) [right of=A1]  {{right{\Large $_3$}}};
      \node[state]           (A3) [right of=A2]  {{right{\Large $_4$}}};
      \node[state]           (B)  [right of=A3]  {{left}};
      \path[->]  (A0) edge [out=60,in=120,looseness=1] node {} (B)  
                      edge              node {} (A1)
                 (A1) edge [out=50,in=130,looseness=1] node {} (B) 
                      edge              node {} (A2)
                 (A2) edge [out=40,in=140,looseness=1] node {} (B) 
                      edge              node {} (A3)
                 (A3) edge [out=30,in=150,looseness=1] node {} (B) 
                 (B)  edge [loop above] node {} (B);
    \end{tikzpicture} \\ (b)
  \end{minipage}
  \caption{(a): The automata describing the correct solution to the reverse problem. 
  The model first has to go to the right while suppressing
  prediction. Then, it has to go to the left and predict what it sees
  at the given moment (this figure illustrates only actions over the
  Input Tape). (b) Another automata that solves the reverse problem
  for short sequences, but does not generalize to arbitrary length
  sequences, unlike (a). Expressive models like LSTMs tend to learn
  such incorrect automata.}
\vspace{-6mm}
  \label{fig:automata} 
\end{figure}
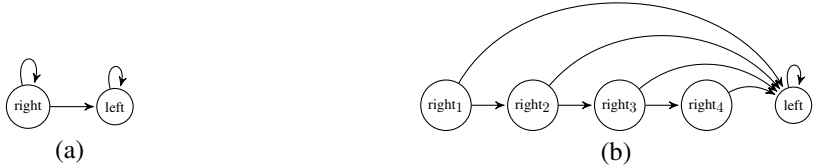

Further insight can be obtained by examining the internal state of the
controller. To do this, we compute the autocorrelation
matrix\footnote{Let $h_i$ be the controller state at time $i$, then
  the autocorrelation $A_{i,j}$ between time-steps $i$ and $j$ is
  given by
  $A_{i,j} = \frac{\langle h_i - E, h_j - E \rangle}{\sigma^2}, i, j =
  1, \dots, T$
  where
  $E = \frac{\sum_{k=1}^T h_k}{T}, \sigma^2 = \frac{\sum_{k=1}^T
    \langle h_k - E, h_k - E\rangle}{T} $.
  $T$ is the number of time steps (i.e. complexity).}  $A$ of the
network state over time when the model is processing a reverse task
example of length $35$, having been trained on sequences of length
$10$ or shorter. For this problem there should be two distinct states:
move right until ``r'' is reached and then move left to the start.
\fig{autocorrelation} plots $A$ for models with three different
controllers. The larger the controller capacity, the less similar the
states are within the two phases of execution, showing how it has not
captured the correct algorithm.  The figure also shows the confidence
in the two actions over time. In the case of the high capacity models,
the initial confidence in the move left action is high, but this drops
off after moving along the sequence. This is because the controller
has learned during training that it should change direction after at
most $10$ steps. Consequently, the unexpectedly long test sequence
makes it unsure of what the correct action is. By contrast, the
simple feed-forward controller does not show this behavior since it is
stateless, thus has no capacity to know where it is within a
sequence. The equivalent automata is shown in \fig{automata}(a), while
\fig{automata}(b) shows the incorrect time-dependent automata learned
by the over-expressive RNN-based controllers. We note that this
argument is empirically supported by our results in \tab{simple}, as
well as related work such as
\cite{graves2014neural} and \cite{joulin2015inferring} which found limited capacity
controllers to be most effective. For example, in the latter case,
the counting and memorization tasks used controllers with just $40$ and
$100$ units respectively. 

\begin{table}[h]
\small
\centering
\includegraphics[width=\linewidth]{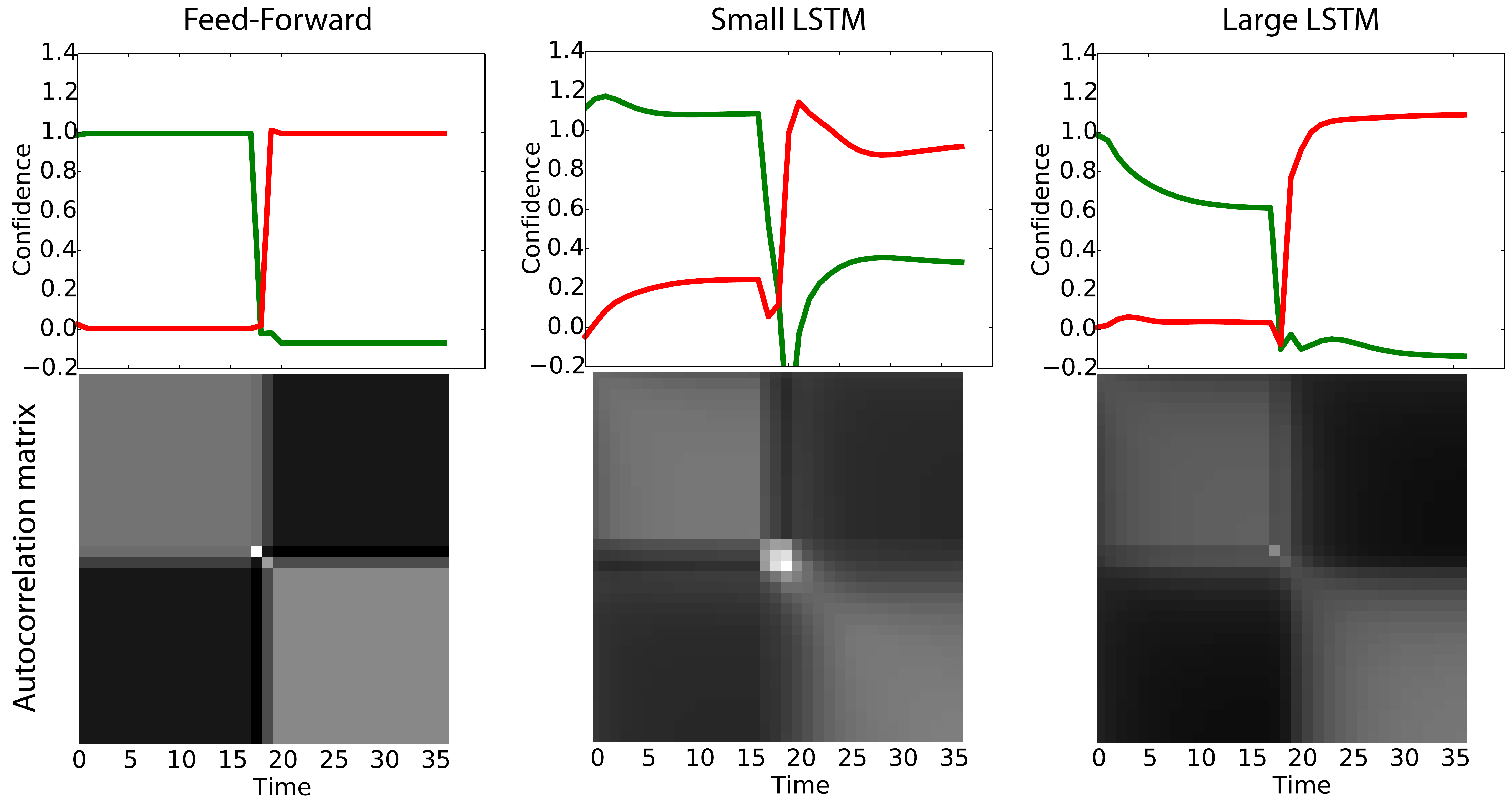} \\
\caption{Three models with different controllers (feed-forward, 200
  unit LSTM and 400 unit LSTM) trained on the reverse task and applied
  to a 20 digit test example. The top row shows confidence values for
  the two actions on the input tape: move left (green) and move right
  (red) as a function of time. The
  correct model should be equivalent to a two-state automata (\fig{automata}), thus we
  expect to see the controller hidden state occupy two distinct
  values. The autocorrelation matrices (whose axes are also time) show this to be the case for
  the feed-forward model -- two distinct blocks of high
  correlation. However, for the LSTM controllers, this structure is
  only loosely present in the matrix, indicating that they have failed
  to learn the correct algorithm. }
\label{fig:autocorrelation}
\vspace{-10mm}
\end{table}

\section{Q-Learning}
\label{sec:qlearn}
In the previous section, we assumed that the optimal controller actions were
given during training. This meant only the output symbols need to be
predicted and these could be learned via backpropagation. 
We now consider the setting where the actions are also learned, to
test the true capabilities of the models to learn simple algorithms
from pairs of input and output sequences.


We use $Q$-learning, a standard reinforcement learning
algorithm to learn a sequence of discrete actions that solves a problem. 
A function $Q$, the estimated sum of future rewards, is
updated during training according to:
\begin{align}
\label{eqn:q}  Q_{t + 1}(s, a) = Q_t(s, a) - \alpha \big[Q_t(s, a) - \big(R(s') +
  \gamma \max_{a}Q_n(s', a)\big)\big]
\end{align}
Taking the action $a$ in state $s$ causes a transition to state $s'$, which in our
case is deterministic.
$R(s')$ is the reward experienced in the state $s'$. The discount
factor is $\gamma$ and $\alpha$ is the learning rate.  
The another commonly considered quantity is $V(s) = \max_a Q(s, a)$. 
$V$ is called the value function, and $V(s)$ is the expected sum of
future rewards starting from the state $s$. Moreover, $Q^*$ and $V^*$ 
are function values for the optimal policy.

Our controller receives a reward of $1$ every time it correctly
predicts a digit (and $0$ otherwise). Since the overall solution to
the task requires all digits to be correct, we terminate a training
episode as soon as an incorrect prediction is made. This learning
environment is {\em non-stationary}, since even if the model initially
picks the right actions, the symbol prediction is unlikely to be
correct, so the model receives no reward. But further on in training,
when the symbol prediction is more reliable, the correct action will
be rewarded\footnote{If we were to use reinforcement to train the
  symbol output as well as the actions, then the environment would be
  stationary. However, this would mean ignoring the reliable signal available
  from direct backpropagation of the symbol output.}. This is important because reinforcement learning algorithms
assume stationarity of the environment, which is not true in our
case. Learning in non-stationary environments is not well
understood and there are no definitive methods to deal with
it. However, empirically we find that this non-stationarity can be
partially addressed by the use of Watkins $Q(\lambda)$
\cite{watkinsq}, as detailed in \secc{watkins}.

\subsection{Dynamic Discount}
\label{sec:time}
The purpose of the reinforcement learning is to learn a policy
that yields the highest sum of the future rewards. $Q$-learning
does it indirectly by learning a $Q$-function. The optimal policy can be
extracted by taking $\argmax$ over $Q(s, \bullet)$. 
Note that shifting or scaling $Q$ induces the same policy. 
We propose to dynamically rescale $Q$ so (i) it is independent 
of the length of the episode and (ii) $Q$ is within a small range,
making it easier to predict.

We define $\hat{Q}$ to be our reparametrization. $\hat{Q}(s, a)$
should be roughly in range $[0, 1]$, and it should correspond to how
close we are to $V^*(s)$.  $Q$ could be decomposed multiplicatively as
${Q(s, a) = \hat{Q}(s,a) V^*(s)}$.  However, in practice, we do not
have access to $V^*(s)$, thus instead we use an estimate of future
rewards based on the total number of digits left in the sequence.
Since every correct prediction yields a reward of $1$, the optimal
policy should achieve sum of future rewards equal to the number of
remaining symbols to predict. The number of remaining symbols to
predict is known and we denote it by $\hat{V}(s)$. Note that this is a
form of supervision, albeit a weak one.

Therefore, we normalize the
$Q$-function by the remaining sum of rewards left in the task:
\begin{align*}
  \hat{Q}(s,a) := \frac{Q(s,a)}{\hat{V}(s)}
\end{align*}
We assume that $s$ transitions to $s'$, and we re-write the $Q$-learning update equations:
\begin{align*}
  \hat{Q}(s, a) &= \frac{R(s')}{\hat{V}(s)} + \gamma \max_{a}\frac{\hat{V}(s')}{\hat{V}(s)} \hat{Q}(s', a) \\
  \hat{Q}_{t + 1}(s, a) &= \hat{Q}_{t}(s, a) -
                          \alpha\big[\hat{Q}_{t}(s, a) -
                          \big(\frac{R(s')}{\hat{V}(s)} + \gamma \max_{a}\frac{\hat{V}(s')}{\hat{V}(s)} \hat{Q}_t(s', a)\big)\big]
\end{align*}
Note that $\hat{V}(s) \geq \hat{V}(s')$, with equality if no digit was
predicted at the current time-step. As the episode progresses, 
the discount factor $\frac{\hat{V}(s')}{\hat{V}(s)}$ decreases,
  making the model greedier. At the end of the sequence, the discount
drops to $\frac{1}{2}$.

\subsection{Watkins $Q(\lambda)$}
\label{sec:watkins}
The update to $Q(s,a)$ in \eqn{q} comes from two parts: the observed
reward $R(s')$ and the estimated future reward $Q(s',a)$.  In our
setting, there are two factors that make the former far more reliable
than the latter: (i) rewards are deterministic and (ii) the
non-stationarity (induced by the ongoing learning of the symbol output
by backpropagation) means that estimates of $Q(s,a)$ are unreliable
as environment evolves. Consequently, the single action recurrence used in
\eqn{q} can be improved upon when on-policy actions are chosen. More precisely, let
$a_t, a_{t+1}, \dots, a_{t+T}$ be consecutive actions induced by $Q$:
\begin{align*}
  a_{t+i} = \argmax_a Q(s_{t+i},a) \\
  s_{t+i} \xrightarrow{a_{t + i}} s_{t + i + 1} 
\end{align*}
Then the optimal $Q^*$ follows the following recursive equation:
\begin{equation*}
Q^*(s_t, a_t) = \sum_{i=1}^T \gamma^{i - 1} R(s_{t + i}) + \gamma^T \max_{a}Q^*(s_{t + n + 1}, a) 
\end{equation*}
and the update rule corresponding to \eqn{q} becomes:
\begin{equation*}
Q_{t + 1}(s_t, a_t) = Q_t(s_t, a_t) - \alpha\big[Q_t(s_t, a_t) -
\big(\sum_{i=1}^T \gamma^{i-1} R(s_{t + i}) + \gamma^T \max_{a}Q_t(s_{t + n + 1}, a)\big)
\big]
\end{equation*}
This is a special form of Watkins $Q(\lambda)$  \cite{watkinsq} where $\lambda=1$.
The classical applications of Watkins $Q(\lambda)$ suggest choosing a small $\lambda$,
which trades-off estimates based on various numbers of future rewards.
$\lambda=0$ rolls back to the classical $Q$-learning. Due to reliability of our rewards,
we found $\lambda = 1$ to be better than $\lambda < 1$, however this
needs further study. 

Note that this unrolling of rewards can only take place until a
non-greedy action is taken. When using an $\epsilon$-greedy policy,
this means we would expect to be able to unroll $\epsilon^{-1}$
steps, on average. For the value of $\epsilon=0.05$ used in our
experiments, this corresponds to $20$ steps on average.
   
\subsection{Penalty on $Q$-function}
\label{sec:penalty}
After reparameterizing the $Q$-function to $\hat{Q}$ (\secc{time}),
the optimal $\hat{Q}^*(s,a)$ should be 1 for the correct action and zero
otherwise. To encourage our estimate $\hat{Q}(s,a)$ to converge to
this, we introduce a penalty that ``pushes down'' on incorrect
actions: $\kappa \| \sum_a \hat{Q}(s,a) - 1 \|^2$.
This has the effect of introducing a margin between correct and
incorrect actions, greatly improving generalization.  We commence
training with $\kappa=0$ and make it non-zero once good accuracy is
reached on short samples (introducing it from the outset hurts
learning). 

\subsection{Reinforcement Learning Experiments}

We apply our enhancements to the six tasks in a series of
experiments designed to examine the contribution of each of
them. Unless otherwise specified, the controller is a 1-layer GRU
model with 200 units. This was selected on the basis of its
mean performance across the six tasks in the supervised setting (see \secc{supervised}). As the performance of reinforcement learning methods tend
to be highly stochastic, we repeat each experiment $10$ times with a
different random seed. Each model is trained using $3
\times 10^7$ characters which takes $\sim4$ hrs. A model is considered
to have successfully solved the task if it able to give a perfect
answer to $50$ test instances, each $100$ digits in length. The GRU
model is trained with a batch size of $20$, a learning rate of
$\alpha=0.1$, using the same initialization as
\cite{glorot2010understanding} but multiplied by 2. 
All tasks are trained with the same curriculum used in the supervised
experiments (and in \cite{joulin2015inferring}), whereby the sequences are initially of complexity $6$
(corresponding to 2 or 3 digits, depending on the task) and once 100\%
accuracy is achieved, increased by $4$ until the model is able to
solve validation sequences of length $100$.  

For 3-row addition, a
more elaborate curriculum was needed which started with examples that did not
involve a carry and contained many zero. The test distribution was
unaffected. Some examples: \qquad$\mborder{\begin{matrix}1\\ 2\\ 2\end{matrix}}$ ; $\mborder{\begin{matrix}2\\ 0\\ 2\end{matrix}}$ ; $\mborder{\begin{matrix}8 & 3\\ 3 & 3\\ 3 & 7\end{matrix}}$ ; $\mborder{\begin{matrix}3 & 2 & 0 & 6 & 9\\ 1 & 3 & 1 & 3 & 1\\ 2 & 8 & 0 & 8 & 3\end{matrix}}$ ; $\mborder{\begin{matrix}8 & 0 & 1 & 8 & 5 & 2 & 0 & 2 & 1\\ 1 & 3 & 1 & 4 & 0 & 7 & 0 & 5 & 4\\ 3 & 1 & 3 & 2 & 7 & 5 & 0 & 7 & 1\end{matrix}}$.

We show results for various combinations of terms in \tab{simple}. The
experiments demonstrate that standard $Q$-learning fails on most of
our tasks (first six columns). Each of our additions (dynamic discount,
Watkins $Q(\lambda)$ and penalty term) give significant
improvements. When all three are used our model is able to succeed at
all tasks, providing the appropriate curriculum and controller are
used. For the reverse and walk tasks, the default GRU controller failed
completely. However, using a feed-forward controller instead enabled
the model to succeed, when dynamic discount and Watkins $Q(\lambda)$
was used. As noted above, the 3-row addition required a more careful
curriculum before the model was able to learn successfully. Increasing
the capacity of the controller (columns 2-4) hurts performance,
echoing \fig{autocorrelation}. The last
two columns of \tab{simple} show results on test sequences of length
1000. Except for multiplication, the models still generalized successfully.

\begin{table}[h]
\scriptsize
\centering
\renewcommand{\arraystretch}{1.1}
\begin{tabular}{|ll||P{0.6cm}|P{0.6cm}||P{0.6cm}|P{0.6cm}|P{0.6cm}|P{0.6cm}|P{0.6cm}|P{0.6cm}||P{0.6cm}|P{0.6cm}|}
\hline 
        & Test length          & 100 & 100 & 100 & 100  & 100  & 100 & 100 & 100 & 1000 & 1000 \\ 
        & \#Units              & 600 & 400 & 200 & 200  & 200  & 200 & 200 & 200 & 200 & 200 \\ 
        & Discount  $\gamma$   & 1  & 1  & 1  & 0.99 & 0.95 & D  & D    &  D    &  D      &    D     \\ 
        & Watkins $Q(\lambda)$ & \no & \no & \no & \no  & \no  & \no & \yes & \yes & \yes  & \yes \\
Task & Penalty                 & \no & \no & \no  & \no  & \no  & \no & \no & \yes & \no    & \yes  \\ \Xhline{4\arrayrulewidth}
Copying &   &           \cellcolor{color_10_7!25} 30\% &  \cellcolor{color_10_4!25} 60\% & \cellcolor{color_10_1!25} 90\% &   \cellcolor{color_10_5!25} 50\%  & \cellcolor{color_10_3!25} 70\%  & \cellcolor{color_10_0!25} 90\% & \cellcolor{color_10_0!25} 100\% & \cellcolor{color_10_0!25} 100\% & \cellcolor{color_10_0!25} 100\% & \cellcolor{color_10_0!25} 100\% \\  \Xhline{2\arrayrulewidth}
\multicolumn{2}{|l||}{Reverse}         & \cellcolor{color_10_10!25} 0\% &  \cellcolor{color_10_10!25} 0\% & \cellcolor{color_10_10!25} 0\% & \cellcolor{color_10_10!25} 0\% & \cellcolor{color_10_10!25} 0\% & \cellcolor{color_10_10!25} 0\% & \cellcolor{color_10_10!25} 0\% & \cellcolor{color_10_10!25} 0\%& \cellcolor{color_10_10!25} 0\% & \cellcolor{color_10_10!25} 0\% \\ \hline
\multicolumn{2}{|l||}{Reverse (FF controller)}         & \cellcolor{color_10_10!25} 0\% &  \cellcolor{color_10_10!25} 0\% & \cellcolor{color_10_10!25} 0\% & \cellcolor{color_10_10!25} 0\% & \cellcolor{color_10_10!25} 0\% & \cellcolor{color_10_10!25} 0\% & \cellcolor{color_10_0!25} 100\% & \cellcolor{color_10_0!25} 90\% & \cellcolor{color_10_0!25} 100\% & \cellcolor{color_10_0!25} 90\% \\ \Xhline{2\arrayrulewidth}
\multicolumn{2}{|l||}{Walk}         &  \cellcolor{color_10_10!25} 0\% & \cellcolor{color_10_10!25} 0\% & \cellcolor{color_10_10!25} 0\% & \cellcolor{color_10_10!25} 0\% & \cellcolor{color_10_10!25} 0\% & \cellcolor{color_10_10!25} 0\% & \cellcolor{color_10_9!25} 10\% & \cellcolor{color_10_1!25} 90\%  & \cellcolor{color_10_9!25} 10\% & \cellcolor{color_10_2!25} 80\% \\ \hline
\multicolumn{2}{|l||}{Walk (FF controller)}         &  \cellcolor{color_10_10!25} 0\% & \cellcolor{color_10_10!25} 0\% & \cellcolor{color_10_10!25} 0\% & \cellcolor{color_10_10!25} 0\% & \cellcolor{color_10_10!25} 0\% & \cellcolor{color_10_10!25} 0\% & \cellcolor{color_10_0!25} 100\% & \cellcolor{color_10_0!25} 100\%  & \cellcolor{color_10_0!25} 100\% & \cellcolor{color_10_0!25} 100\% \\ \Xhline{2\arrayrulewidth}
2-row Addition &          & \cellcolor{color_10_9!25} 10\% &  \cellcolor{color_10_3!25} 70\% & \cellcolor{color_10_4!25} 70\% &   \cellcolor{color_10_3!25} 70\%  & \cellcolor{color_10_2!25} 80\%  &  \cellcolor{color_10_4!25} 60\% &  \cellcolor{color_10_4!25} 60\% & \cellcolor{color_10_0!25} 100\% &  \cellcolor{color_10_6!25} 40\% & \cellcolor{color_10_0!25} 100\% \\  \Xhline{2\arrayrulewidth}
\multicolumn{2}{|l||}{3-row Addition}         & \cellcolor{color_10_10!25} 0\% & \cellcolor{color_10_10!25} 0\% & \cellcolor{color_10_10!25} 0\% & \cellcolor{color_10_10!25} 0\% & \cellcolor{color_10_10!25} 0\% & \cellcolor{color_10_10!25} 0\% & \cellcolor{color_10_10!25} 0\% & \cellcolor{color_10_10!25} 0\%& \cellcolor{color_10_10!25} 0\% & \cellcolor{color_10_10!25} 0\% \\ \hline
\multicolumn{2}{|l||}{3-row Addition (extra curriculum)}          &  \cellcolor{color_10_10!25} 0\% & \cellcolor{color_10_5!25} 50\% & \cellcolor{color_10_1!25} 80\%  & \cellcolor{color_10_6!25} 40\% &  \cellcolor{color_10_5!25} 50\% &  \cellcolor{color_10_5!25} 50\% &  \cellcolor{color_10_1!25} 80\%  &  \cellcolor{color_10_1!25} 80\%  & \cellcolor{color_10_9!25} 10\% &  \cellcolor{color_10_4!25} 60\%\\ \Xhline{2\arrayrulewidth}
\multicolumn{2}{|l||}{Single Digit Multiplication}   & \cellcolor{color_10_10!25} 0\% & \cellcolor{color_10_10!25} 0\% & \cellcolor{color_10_10!25} 0\% & \cellcolor{color_10_10!25} 0\% & \cellcolor{color_10_10!25} 0\% & \cellcolor{color_10_0!25} 100\% & \cellcolor{color_10_0!25} 100\% & \cellcolor{color_10_0!25} 100\% & \cellcolor{color_10_10!25} 0\% & \cellcolor{color_10_10!25} 0\% \\
\hline
\end{tabular}
\caption{Success rates for classical $Q$-learning (columns 2-5) versus our enhanced $Q$-learning. A GRU-based controller is
  used on all tasks, except reverse and walk which use a feed-forward
  network. Curriculum learning was also used for the 3-row addition
  task (see text for details). When dynamic discount (D), Watkins
  $Q(\lambda)$ and the penalty term are all used the model
  consistently succeeds on all tasks. The model still performs well on
  test sequences of length 1000, apart from the multiplication
  task. Increasing the capacity of the controller results in worse
  performance (columns 2-4).}
\label{tab:simple}
\end{table}

\begin{figure}[h]
  \centering
  \includegraphics[width=\linewidth]{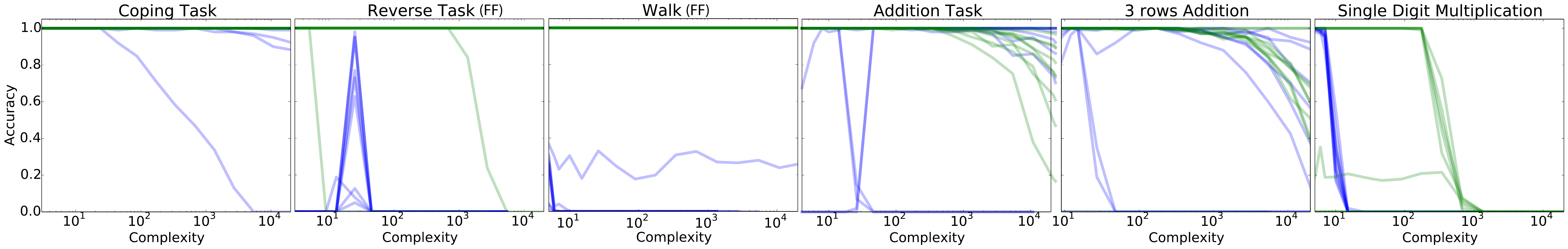}
 \caption{Test accuracy as a function of task complexity ($10$
   runs) for standard $Q$-learning (blue) and our 
   enhanced version (dynamic discount, Watkins $Q(\lambda)$ and
   penalty term). Accuracy corresponds to the fraction of correct test cases
   (all digits must be predicted correctly for the instance to be
   considered correct). }
  \label{fig:accuracy}
\end{figure}

\fig{accuracy} shows accuracy as a function of test example complexity
for standard $Q$-learning and our enhanced version. The difference is
performance is clear. At very high complexity, corresponding to
$1000$'s of digits, the accuracy starts to drop on the more
complicated tasks. We note that these trends are essentially the same
as those observed in the supervised setting (\fig{supervised}),
suggesting that $Q$-learning is not to blame. Instead, the inability
of the controller to learn an automata seems to be the
cause. Potential solutions to this might include (i) noise injection,
(ii) discretization of state, (iii) a state error correction
mechanism or (iv) regularizing the learned automata using MDL principles. However, this issue, the inability of RNN to perfectly represent
an automata can be examined separately from the setting
where actions have to be learnt (i.e.~in the supervised domain).

Further results can be found in the appendices. 
For the addition task, our model was able to discover multiple correct
solutions, each with a different movement pattern over the input tape
(see Appendix A). \tab{base} in Appendix B sheds light on the trade-off
between errors in actions and errors in symbol prediction by varying
the base used in the arithmetic operations and hence the size of the target vocabulary.
Appendix C explores the use of non-integer rewards. Surprisingly, this
slows down training, relative to the 0/1 reward structure.

\section{Discussion}

We have explored the ability of neural network models to learn
algorithms for simple arithmetic operations. Through experiments with
supervision and reinforcement learning, we have shown that they are
able to do this successfully, albeit with caveats. $Q$-learning was
shown to work as well as the supervised case. But, disappointingly, we
were not able to find a single controller that could solve all
tasks. We found that for some tasks, generalization ability was sensitive to
the memory capacity of the controller: too little and it would be
unable to solve more complex tasks that rely on carrying state across
time; too much and the resulting model would overfit the length of the
training sequences. Finding automatic methods to control model
capacity would seem to be important in developing robust models for
this type of learning problem.

\subsubsection*{Acknowledgments}
We wish to thank Jason Weston, Marc'Aurelio Ranzato and Przemys\l{}aw Mazur for useful discussions, and comments.
We also thank Christopher Olah for LSTM figures that have been used in the paper and the accompanying video.

\bibliography{bibliography}

\begin{thebibliography}{20}
\providecommand{\natexlab}[1]{#1}
\providecommand{\url}[1]{\texttt{#1}}
\expandafter\ifx\csname urlstyle\endcsname\relax
  \providecommand{\doi}[1]{doi: #1}\else
  \providecommand{\doi}{doi: \begingroup \urlstyle{rm}\Url}\fi

\bibitem[Cho et~al.(2014)Cho, van Merrienboer, Gulcehre, Bougares, Schwenk, and
  Bengio]{cho2014learning}
Cho, Kyunghyun, van Merrienboer, Bart, Gulcehre, Caglar, Bougares, Fethi,
  Schwenk, Holger, and Bengio, Yoshua.
\newblock Learning phrase representations using rnn encoder-decoder for
  statistical machine translation.
\newblock \emph{arXiv preprint arXiv:1406.1078}, 2014.

\bibitem[Das et~al.(1992)Das, Giles, and Sun]{Das92}
Das, Sreerupa, Giles, C~Lee, and Sun, Guo-Zheng.
\newblock Learning context-free grammars: Capabilities and limitations of a
  recurrent neural network with an external stack memory.
\newblock In \emph{In Proceedings of The Fourteenth Annual Conference of
  Cognitive Science Society}, 1992.

\bibitem[Glorot \& Bengio(2010)Glorot and Bengio]{glorot2010understanding}
Glorot, Xavier and Bengio, Yoshua.
\newblock Understanding the difficulty of training deep feedforward neural
  networks.
\newblock In \emph{International conference on artificial intelligence and
  statistics}, pp.\  249--256, 2010.

\bibitem[Goldberg(1989)]{Goldberg}
Goldberg, David~E.
\newblock \emph{Genetic Algorithms in Search, Optimization and Machine
  Learning}.
\newblock Addison-Wesley Longman Publishing Co., Inc., Boston, MA, USA, 1st
  edition, 1989.
\newblock ISBN 0201157675.

\bibitem[Graves et~al.(2014)Graves, Wayne, and Danihelka]{graves2014neural}
Graves, Alex, Wayne, Greg, and Danihelka, Ivo.
\newblock Neural turing machines.
\newblock \emph{arXiv preprint arXiv:1410.5401}, 2014.

\bibitem[Grefenstette et~al.(2015)Grefenstette, Hermann, Suleyman, and
  Blunsom]{grefenstette2015learning}
Grefenstette, Edward, Hermann, Karl~Moritz, Suleyman, Mustafa, and Blunsom,
  Phil.
\newblock Learning to transduce with unbounded memory.
\newblock \emph{arXiv preprint arXiv:1506.02516}, 2015.

\bibitem[Hasselt(2010)]{hasselt2010double}
Hasselt, Hado~V.
\newblock Double q-learning.
\newblock In \emph{Advances in Neural Information Processing Systems}, pp.\
  2613--2621, 2010.

\bibitem[Hochreiter \& Schmidhuber(1997)Hochreiter and
  Schmidhuber]{hochreiter1997long}
Hochreiter, Sepp and Schmidhuber, J{\"u}rgen.
\newblock Long short-term memory.
\newblock \emph{Neural computation}, 9\penalty0 (8):\penalty0 1735--1780, 1997.

\bibitem[Holland(1992)]{Holland}
Holland, John~H.
\newblock \emph{Adaptation in Natural and Artificial Systems: An Introductory
  Analysis with Applications to Biology, Control and Artificial Intelligence}.
\newblock MIT Press, Cambridge, MA, USA, 1992.
\newblock ISBN 0262082136.

\bibitem[Joulin \& Mikolov(2015)Joulin and Mikolov]{joulin2015inferring}
Joulin, Armand and Mikolov, Tomas.
\newblock Inferring algorithmic patterns with stack-augmented recurrent nets.
\newblock \emph{arXiv preprint arXiv:1503.01007}, 2015.

\bibitem[Liang et~al.(2013)Liang, Jordan, and Klein]{liang2013learning}
Liang, Percy, Jordan, Michael~I, and Klein, Dan.
\newblock Learning dependency-based compositional semantics.
\newblock \emph{Computational Linguistics}, 39\penalty0 (2):\penalty0 389--446,
  2013.

\bibitem[Mnih et~al.(2013)Mnih, Kavukcuoglu, Silver, Graves, Antonoglou,
  Wierstra, and Riedmiller]{mnih2013playing}
Mnih, Volodymyr, Kavukcuoglu, Koray, Silver, David, Graves, Alex, Antonoglou,
  Ioannis, Wierstra, Daan, and Riedmiller, Martin.
\newblock Playing atari with deep reinforcement learning.
\newblock \emph{arXiv preprint arXiv:1312.5602}, 2013.

\bibitem[Nordin(1997)]{nordin1997evolutionary}
Nordin, Peter.
\newblock \emph{Evolutionary program induction of binary machine code and its
  applications}.
\newblock Krehl Munster, 1997.

\bibitem[Schmidhuber(2004)]{schmidhuber2004optimal}
Schmidhuber, J{\"u}rgen.
\newblock Optimal ordered problem solver.
\newblock \emph{Machine Learning}, 54\penalty0 (3):\penalty0 211--254, 2004.

\bibitem[Solomonoff(1964)]{solomonoff1964formal}
Solomonoff, Ray~J.
\newblock A formal theory of inductive inference. {Part I}.
\newblock \emph{Information and control}, 7\penalty0 (1):\penalty0 1--22, 1964.

\bibitem[Sukhbaatar et~al.(2015)Sukhbaatar, Szlam, Weston, and
  Fergus]{sukhbaatar2015weakly}
Sukhbaatar, Sainbayar, Szlam, Arthur, Weston, Jason, and Fergus, Rob.
\newblock Weakly supervised memory networks.
\newblock \emph{arXiv preprint arXiv:1503.08895}, 2015.

\bibitem[Sutton \& Barto(1998)Sutton and Barto]{sutton1998reinforcement}
Sutton, Richard~S and Barto, Andrew~G.
\newblock \emph{Reinforcement learning: An introduction}, volume~1.
\newblock MIT press Cambridge, 1998.

\bibitem[Watkins(1989)]{watkinsq}
Watkins, Chris.
\newblock \emph{Learning from Delayed Rewards}.
\newblock PhD thesis, Cambrdige University, 1989.

\bibitem[Wineberg \& Oppacher(1994)Wineberg and
  Oppacher]{wineberg1994representation}
Wineberg, Mark and Oppacher, Franz.
\newblock A representation scheme to perform program induction in a canonical
  genetic algorithm.
\newblock In \emph{Parallel Problem Solving from Nature—PPSN III}, pp.\
  291--301. Springer, 1994.

\bibitem[Zaremba \& Sutskever(2015)Zaremba and
  Sutskever]{zaremba2015reinforcement}
Zaremba, Wojciech and Sutskever, Ilya.
\newblock Reinforcement learning neural turing machines.
\newblock \emph{arXiv preprint arXiv:1505.00521}, 2015.

\end{thebibliography}
\bibliographystyle{iclr2016_conference}

\newpage

\section*{Appendix A: Different Solutions to Addition Task }
\begin{figure}[h]
  \begin{center}
    \begin{picture}(280, 100)
      \multiput(0,50)(20,0){3}{
        \put(0, 0){\framebox(20, 20){}}
        \put(0, 20){\framebox(20, 20){}}
        \put(20, 0){\framebox(20, 20){}}
        \put(20, 20){\framebox(20, 20){}}

        \put(34, 30){\vector(-0.2, -1){4}}
        \put(29, 10){\vector(-0.2, 1){4}}
        \put(25, 30){\vector(-1, 0){12}}
      }
      \multiput(100, 50)(20,0){3}{
        \put(0, 0){\framebox(20, 20){}}
        \put(0, 20){\framebox(20, 20){}}
        \put(20, 0){\framebox(20, 20){}}
        \put(20, 20){\framebox(20, 20){}}

        \put(29, 30){\vector(-0.2, -1){4}}
        \put(14, 10){\vector(-0.2, 1){4}}
        \put(25, 10){\vector(-1, 0){12}}
      }
      \multiput(200, 50)(40,0){2}{
        \put(0, 0){\framebox(20, 20){}}
        \put(0, 20){\framebox(20, 20){}}
        \put(20, 0){\framebox(20, 20){}}
        \put(20, 20){\framebox(20, 20){}}

        \put(30, 30){\vector(0, -1){20}}
        \put(30, 10){\vector(-1, 0){20}}
        \put(10, 10){\vector(0, 1){20}}
        \put(10, 30){\vector(-1, 0){20}}
      }
    \end{picture}
  \end{center}

  \caption{Our model found three different solutions to the addition
    task, all of which give the correct answer. The arrows show the
    trajectory of the read head over the input grid.}
  \label{fig:addition_solutions}
\end{figure}
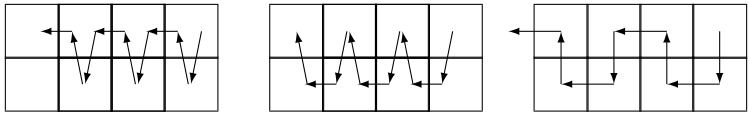

On examination of the models learned on the addition task, we notice
that three different solutions were discovered. While they all give the correct
answer, they differ in their actions over the input
grid, as shown in \fig{addition_solutions}.

\section*{Appendix B:Reward Frequency vs Reward Reliability}

We explore how learning time varies as the size of the target
vocabulary is varied. This trades off reward frequency and
reliability. For small vocabularies, the reward occurs more often but
is less reliable since the chance of the wrong action sequence
yielding the correct result is relatively high (and vice-versa for for
larger vocabularies). For copying and reverse tasks, altering the
vocabulary size just alters the variety of symbols on the
tape. However, for the arithmetic operations this involves a change of
base, which influences the task in a more complex way. For instance,
addition in base $4$ requires the memorization of  digit-to-digit addition
table of size $16$ instead of $100$ for the base $10$. \tab{base}
shows the median training time as a function of vocabulary size. The
results suggest that an infrequent but reliable reward is preferred to
a frequent but noisy one.

\begin{table}[h]
\tiny
\centering
\renewcommand{\arraystretch}{1.15}
\begin{tabular}{|lccccccccc|}
\hline 
\multicolumn{1}{|l|}{\backslashbox{Task}{Vocabulary Size}} & \multicolumn{1}{|c|}{2} & \multicolumn{1}{|c|}{3} & \multicolumn{1}{|c|}{4} & \multicolumn{1}{|c|}{5} & \multicolumn{1}{|c|}{6} & \multicolumn{1}{|c|}{7} & \multicolumn{1}{|c|}{8} & \multicolumn{1}{|c|}{9} & \multicolumn{1}{|c|}{10}\\\hline 
Coping & \cellcolor{color_10_6!25} 1.4 & \cellcolor{color_10_4!25} 0.6 & \cellcolor{color_10_4!25} 0.6 & \cellcolor{color_10_4!25} 0.6 & \cellcolor{color_10_4!25} 0.6 & \cellcolor{color_10_4!25} 0.5 & \cellcolor{color_10_4!25} 0.6 & \cellcolor{color_10_4!25} 0.6 & \cellcolor{color_10_4!25} 0.6\\
Reverse (FF controller) & \cellcolor{color_10_5!25} 6.5 & \cellcolor{color_10_10!25} 23.8 & \cellcolor{color_10_4!25} 3.1 & \cellcolor{color_10_4!25} 3.6 & \cellcolor{color_10_4!25} 3.8 & \cellcolor{color_10_4!25} 2.5 & \cellcolor{color_10_4!25} 2.8 & \cellcolor{color_10_3!25} 2.0 & \cellcolor{color_10_4!25} 3.1\\
Walk (FF controller) & \cellcolor{color_10_7!25} 8.7 & \cellcolor{color_10_5!25} 6.9 & \cellcolor{color_10_5!25} 6.8 & \cellcolor{color_10_2!25} 4.0 & \cellcolor{color_10_4!25} 6.2 & \cellcolor{color_10_3!25} 5.3 & \cellcolor{color_10_2!25} 4.4 & \cellcolor{color_10_2!25} 3.9 & \cellcolor{color_10_10!25} 11.1\\
Addition & \cellcolor{color_10_10!25} 250.0 & \cellcolor{color_10_4!25} 30.9 & \cellcolor{color_10_4!25} 14.5 & \cellcolor{color_10_4!25} 26.1 & \cellcolor{color_10_4!25} 21.8 & \cellcolor{color_10_4!25} 21.9 & \cellcolor{color_10_4!25} 25.0 & \cellcolor{color_10_4!25} 23.4 & \cellcolor{color_10_4!25} 21.1\\
3-number Addition (extra curriculum) & \cellcolor{color_10_10!25} 250.0 & \cellcolor{color_10_0!25} 61.5 & \cellcolor{color_10_10!25} 250.0 & \cellcolor{color_10_10!25} 250.0 & \cellcolor{color_10_3!25} 112.2 & \cellcolor{color_10_6!25} 178.2 & \cellcolor{color_10_2!25} 93.8 & \cellcolor{color_10_1!25} 79.1 & \cellcolor{color_10_1!25} 81.9\\
Single Digit Multiplication & invalid & \cellcolor{color_10_0!25} 6.2 & \cellcolor{color_10_4!25} 17.8 & \cellcolor{color_10_5!25} 20.9 & \cellcolor{color_10_5!25} 21.4 & \cellcolor{color_10_5!25} 21.5 & \cellcolor{color_10_6!25} 22.3 & \cellcolor{color_10_6!25} 23.3 & \cellcolor{color_10_7!25} 24.7\\
\hline 
\end{tabular}
\caption{Median training time (minutes) over $10$ runs as we vary the
  base used (hence vocabulary size) on different problems. Training
  stops when the model successfully generalizes to test sequences of
  length $100$. The results show
  the relative importance of reward frequency versus reliability, with
  the latter being more important. }
\label{tab:base}
\end{table}

\section*{Appendix C: Reward Structure}
Reward in reinforcement learning systems drives the learning process. 
In our setting we control the rewards, deciding when, and how much to give.
We now examine various kinds of rewards and their influence on the learning time
of our system. 

Our vanilla setting gives a reward of $1$ for every correct
prediction, and reward $0$ for every incorrect one. We refer to this
setting as ``0/1 reward''. We consider two other settings in addition
to this, both of which rely on the probabilities of the correct
prediction.  Let $y$ be the target symbol and
$p_i = p(y = i), i \in [0, 9]$ be the probability of predicting label
$i$.

In setting ``Discretized reward'', we sort $p_i$. 
That gives us an order on indices $a_1, a_2, \dots, a_{10}$,
i.e. $p_{a_1} \geq p_{a_2} \geq p_{a_3} \dots \geq p_{a_{10}}$.
``Discretized reward'' yields reward $1$ iff $a_1 \equiv y$, reward $\frac{1}{2}$ iff $a_2 \equiv y$,
and reward $\frac{1}{3}$ iff $a_3 \equiv y$. Otherwise, environment gives a reward $0$.
In the ``Continuous reward'' setting, a reward of $p_y$ is given for every prediction.
One could also consider reward $\log(p_y)$, however this quantity is unbounded,
and further processing might be necessary to make it work.

Table \ref{tab:more_reward} gives results for the three different
reward structures, showing training time for the five tasks (training
is stopped once the model generalizes to test sequences of length
$100$).  One might expect that a continuous reward would convey more
information than a discrete one, thus result in faster training.
However, the results do not support this hypothesis, as training seems
harder with continuous reward than a discrete one. We hypothesize,
that the continuous reward makes environment less stationary, which
might make $Q$-learning less efficient, although this needs further verification.

\begin{table}[h]
\tiny
\centering
\renewcommand{\arraystretch}{1.15}
\begin{tabular}{|lccc|}
\hline 
\multicolumn{1}{|l|}{\backslashbox{Task}{Reward Type}} & \multicolumn{1}{|c|}{0/1 reward} & \multicolumn{1}{|c|}{Discretized reward} & \multicolumn{1}{|c|}{Continuous reward}\\\hline 
Coping & \cellcolor{color_10_4!25} 0.6 & \cellcolor{color_10_4!25} 0.6 & \cellcolor{color_10_5!25} 0.8\\
Reverse (FF controller) & \cellcolor{color_10_2!25} 3.1 & \cellcolor{color_10_2!25} 3.1 & \cellcolor{color_10_10!25} 59.7\\
Walk (FF controller) & \cellcolor{color_10_2!25} 11.1 & \cellcolor{color_10_2!25} 9.5 & \cellcolor{color_10_10!25} 250.0\\
Addition & \cellcolor{color_10_2!25} 21.1 & \cellcolor{color_10_3!25} 21.6 & \cellcolor{color_10_9!25} 24.2\\
3-number Addition (extra curriculum) & \cellcolor{color_10_2!25} 81.9 & \cellcolor{color_10_2!25} 77.9 & \cellcolor{color_10_10!25} 131.9\\
Single Digit Multiplication & \cellcolor{color_10_1!25} 24.7 & \cellcolor{color_10_6!25} 26.5 & \cellcolor{color_10_6!25} 26.6\\
\hline
\end{tabular}
\caption{Median training time (minutes) for the five tasks for the
  three different reward structures.
  ``0/1 reward'': the model gets a reward of $1$ for every correct prediction, and $0$ otherwise. 
  ``Discretized reward'' provides a few more values of reward
  prediction, if sufficiently close to the
  correct one. ``Continuous reward'' gives a probability of correct
  answer as the reward. See text for details.}
\label{tab:more_reward}
\end{table}

\end{document}